# WEIGHT INITIALIZATION TECHNIQUES FOR DEEP LEARNING ALGORITHMS IN REMOTE SENSING: RECENT TRENDS AND FUTURE PERSPECTIVES

*Wadii Boulila[1,2], Maha Driss[1,2], Mohamed Al-Sarem[2], Faisal Saeed[2], Moez Krichen[3,4]*

[1]RIADI Laboratory, National School of Computer Sciences, University of Manouba, Tunisia
[2]IS Department, College of Computer Science and Engineering, Taibah University, Saudi Arabia
[3]CS Department, Faculty of CSIT, Al-Baha University, Saudi Arabia
[4]ReDCAD Laboratory, University of Sfax, Tunisia

**ABSTRACT**

During the last decade, several research works have focused on providing novel deep learning methods in many application fields. However, few of them have investigated the weight initialization process for deep learning, although its importance is revealed in improving deep learning performance. This can be justified by the technical difficulties in proposing new techniques for this promising research field. In this paper, a survey related to weight initialization techniques for deep algorithms in remote sensing is conducted. This survey will help practitioners to drive further research in this promising field. To the best of our knowledge, this paper constitutes the first survey focusing on weight initialization for deep learning models.

*Index Terms*— Review, deep learning, weight initialization, remote sensing

## 1. INTRODUCTION

Deep learning (DL) is one of the most promising machine learning techniques that has been applied in several domains. In this paper, we focus on the application of DL in remote sensing (RS). DL has been applied in many RS-related applications such as crop yield tracking, land cover change monitoring, disaster prediction, and urban planning, and it demonstrates excellent results as demonstrated in [1-6]. DL-based architectures are characterized by many hidden layers of neurons. However, the main limitation of these architectures is the long time required for training. Obtaining excellent accuracy and reasonable training time is a challenging objective for the DL research community. Selecting an appropriate weight initialization strategy is critical when training DL techniques. Weight initialization represents the manner of setting initial weight values of a neural network layer. According to [7], DL methods are very sensitive to the values of initial weights. Initialization of weight seeks to assist in establishing a stable neural network learning bias and shorten convergence time.

The main motivation behind this paper is that many researchers have drawn great attention to developing new DL algorithms, however, few of them have focused on proposing a new method for weight initialization, especially in RS. In this paper, we provide a review of the literature regarding weight initialization in RS, which will help practitioners to drive further research in this promising field.

The remainder of this paper is organized as follows. In Section 2, we briefly detail the main weight initialization techniques. In Section 3, existing research works related to weight initialization techniques in RS are reviewed. Lessons learned and future perspectives are presented in Section 4.

## 2. WEIGHT INITIALIZATION TECHNIQUES

The goal of weight initialization is to prevent layer activation outputs from exploding or vanishing during the training of the DL technique. Training the network without a useful weight initialization can lead to a very slow convergence or an inability to converge [8].

Figure 1 depicts the process of weight initialization for a given layer of the DL network.

$$W = \begin{bmatrix} W_{11} & W_{12} & \cdots & W_{1j} \\ W_{21} & W_{22} & \cdots & W_{2j} \\ \vdots & \vdots & \cdots & \vdots \\ W_{n1} & W_{n2} & \cdots & W_{nj} \end{bmatrix}$$

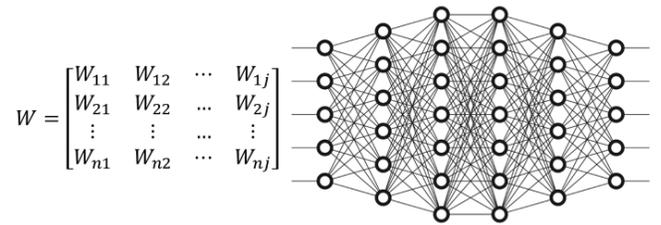

Fig. 1. Weight initialization process.

The most used weight initialization techniques are described as follows [9]:

*a. All-zeros initialization and Constant initialization*

This method sets all weights to zeros (respectively to constant). Also, all activations in all neurons are the same, and therefore all calculations are the same, making which makes the concerned model a linear model.

*b. Random initialization*

This technique improves the symmetry-breaking process and provides much greater precision. The weights are initialized very near to zero and randomly. This method prevents from learning the same feature for input parameters.

*c. LeCun initialization : normalize variance*

LeCun initialization aims to prevent the vanishing or explosion of the gradients during the backpropagation by solving the growing variance with the number of inputs and by setting constant variance.

*d. Xavier initialization (Glorot initialization)*

Xavier proposed a more straightforward method, where the weights such as the variance of the activations are the same across every layer. This will prevent the gradient from exploding or vanishing.

*e. He initialization (Kaiming initialization)*

This initialization preserves the non-linearity of activation functions such as ReLU activations. Using the He method, we can reduce or magnify the magnitudes of inputs exponentially.

TABLE I. COMPARISON BETWEEN WEIGHT INITIALIZATION TECHNIQUES

| Initialization method | Pros. | Cons. |
|---|---|---|
| All-zeros / constant | Simplicity | Symmetry problem leading neurons to learn the same features |
| Random | Improves the symmetry-breaking process | - A saturation may occur leading to a vanishing gradient<br>- The slope or gradient is small, which can cause the gradient descent to be slow |
| LeCun | Solves growing variance and gradient problems | - Not useful in constant-width networks<br>- Takes into account the forward propagation of the input signal<br>- This method is not useful when the activation function is non-differentiable |
| Xavier | Decreases the probability of the gradient vanishing/exploding problem | - This method is not useful when the activation function is non-differentiable<br>- Dying neuron problem during the training |
| He | Solves dying neuron problems | - This method is not useful for layers with differentiable activation function such as ReLU or LeakyReLU |

In Table I, a comparison between weight initialization techniques is depicted. In this table, we compared the main weight initialization techniques according to their advantages and limitations.

## 3. WEIGHT INITIALIZATION TECHNIQUES IN RS

In this section, we focus on research works that investigated weight initialization techniques in RS (Table II). In this survey, we reviewed 17 relevant research papers according to the following criteria: application case study, DL model, and weight initialization method.

TABLE II. RESEARCH WORKS INVESTIGATING WEIGHT INITIALIZATION IN RS DOMAIN

| Reference | Application case study | DL model | Weight initialization method |
|---|---|---|---|
| [10] | Semantic segmentation of small objects | CNN | He |
| [11] | Classification of hyperspectral image | CDL | Auto-encoder |
| [12] | Automatic target recognition in synthetic aperture radar | CNN | Random distribution |
| [13] | Single image super-resolution | CNN | He |
| [14] | Surface water mapping | CNN | He |
| [15] | Classification of oceanic eddies | CNN | Gaussian distribution |
| [16] | Hyperspectral image classification | CNN | MSRA (for Microsoft Research Asia) |
| [17] | Infrastructure Quality Assessment | CNN | Xavier |
| [18] | Remote sensing image fusion | CNN | He |
| [19] | Polarimetric synthetic aperture radar image classification | Complex-valued deep fully CNN | A new method is proposed |
| [20] | Hyperspectral image classification | CNN | Random distribution |
| [21] | Multi-label aerial image classification | CNN BiLSTM | Xavier |
| [22] | Building change detection | DBN ELM | Random distribution |
| [23] | Multispectral image classification | RNN LSTM | Xavier |
| [24] | Rice lodging canopy | CNN | Xavier |
| [25] | Road extraction | PSNet | A new method is proposed |
| [26] | Land cover change detection | LSTM CNN | Random distribution |

BiLSTM: Bidirectional Long Short-Term Memory
CNN: Convolutional Neural Network
CDL: Contextual Deep Learning
DBN: Deep Belief Network
ELM: Extreme Learning Machine
LSTM: Long Short-Term Memory
PSNet: Parallel Softplus Network

From Table II, we note that weight initialization is a very recent topic in RS (most research works that are cited in Table

II are conducted after 2016). Besides, these research works did not focus on developing weight initialization but use built-in techniques such as random, Xavier, and He. Only two papers have proposed a new technique for weight initialization. Another important observation is that the CNN model is used in most of the articles we studied in this survey (13 articles out of 17).

## 4. LESSON LEARNED AND FUTURE PERSPECTIVES

### 4.1. Lessons learned

According to several studies reported in the literature review, it is important to use weight initialization techniques for complex datasets [9]. Weight initialization has an important role in the training of complex data, especially when working with heuristic-based methods, which are designed by using certain properties of activation functions. In the case of a small weight initialization, the input of neurons will be small leading to the loss of the non-linearity of the activation function. Otherwise, in the case of a large weight initialization, the input of neurons will be large leading to a saturation of the activation function. Choosing the appropriate weight initializers will help to obtain better performance for the DL model [27]. Also, a good initialization of weights helps gradient-based methods to quickly converge.

According to the literature, selecting the appropriate weight initialization method is an open issue. However, in this paper, we will provide some tips that can help researchers in this issue. Indeed, LeCun and Xavier initialization methods provide good results in the case of differentiable activation function such as sigmoid. The He initialization method provides good results in the case of non-differentiable activation functions such as ReLU. In most cases, DNN models are based on ReLU activation function. Hence, it is better to use the He method for weight initialization in the case of DNN.

### 4.2. Future perspectives

After studying literature related to weight initialization techniques in RS, we highlight, in this subsection, future perspectives, which can be used for further research studies in this promising field. These future perspectives are summarized in the following points:

- *Mathematical explanations of weight initialization strategies*: there is a lack of mathematical models that help to determine the most appropriate weight initialization method for a given dataset and DL model [9].
- *Big RS data context*: big RS data provide a great opportunity for DL to extract meaningful insights for several case studies. However, the massive volume of data (complexity, variability, and velocity) is a challenging task to DL techniques. Training DL techniques over this type of data is a daunting process, requiring computationally expensive processing, which is unfeasible for large networks. Investigating the weight initialization for this type of data remains a very promising research field [27].
- *Uncertainty and missing data*: working with weight initialization in the context of missing labels in the training dataset is a challenging task. Several works suggest using autoencoders as a weight initialization method to assign weights to each layer of the model [28] [29].

## 5. CONCLUSION

Weight initialization plays an important role in improving the training process of DL methods. In this paper, we have reviewed weight initialization techniques for DL algorithms in RS according to 3 different criteria: the application case study, the used DL model, and the used weight initialization method. After that, we highlighted lessons learned and future perspectives to drive further research in this promising field. According to our review of the literature, we note that most existing research works rely on existing weight initialization methods, which require further optimization and contextualization in terms of the used datasets and DL models. Nowadays, with the new progress made on data acquisition, we are facing massive RS data, which requires more investigation in the weight initialization research field [30-33].